\documentclass[letterpaper, 10 pt, conference]{ieeeconf}

\IEEEoverridecommandlockouts                              

\usepackage{cite}
\usepackage{amsmath,amssymb,amsfonts,multirow,tabularx,lipsum,algorithm,algpseudocode}
\usepackage{graphicx}
\usepackage{textcomp}
\usepackage{xcolor}
\usepackage{lipsum}

\renewcommand{\baselinestretch}{0.96}

\begin{document}

\title{Zero-shot Active Visual Search (ZAVIS): \\
Intelligent Object Search for Robotic Assistants
}

\author{Jeongeun Park$^{1}$, 
Taerim Yoon$^{1}$, Jejoon Hong$^{2}$, Youngjae Yu$^{3}$, Matthew Pan$^{4}$,
and Sungjoon Choi$^{1}$
\thanks{$^{1}$Jeongeun Park, Taerim Yoon, and Sungjoon Choi are with 
the Department of Artificial Intelligence, 
Korea University, Seoul, Korea
{\tt\footnotesize 
\{baro0906,taerimyoon,sungjoon-choi\}@korea.ac.kr}.}%
\thanks{$^{2} $Jejoon Hong is with 
the Department of Mechanical Engineering, 
Korea University, Seoul, Korea
{\tt\footnotesize hdavid0510@korea.ac.kr}.}
\thanks{$^{3} $ Youngjae Yu is with 
Allen Institute for AI, 
{\tt\footnotesize youngjaey@allenai.org }.}
\thanks{$^{4} $Matthew Pan is with 
the Department of Electrical and Computer Engineering, Queens University, Kingston, Canada
{\tt\footnotesize matthew.pan@queensu.ca}.}%
}

\maketitle
\begin{abstract}
In this paper, we focus on the problem of efficiently locating a target object described with \textit{free-form language} using a mobile robot equipped with vision sensors (e.g., an RGBD camera). Conventional active visual search predefines a set of objects to search for, rendering these techniques restrictive in practice. To provide added flexibility in active visual searching, we propose a system where a user can enter target commands using free-form language; we call this system Zero-shot Active Visual Search (ZAVIS). ZAVIS detects and plans to search for a target object inputted by a user through a semantic grid map represented by static landmarks (e.g., desk or bed). For efficient planning of object search patterns, ZAVIS considers commonsense knowledge-based co-occurrence and predictive uncertainty while deciding which landmarks to visit first. We validate the proposed method with respect to SR (success rate) and SPL (success weighted by path length) in both simulated and real-world environments. The proposed method outperforms previous methods in terms of SPL in simulated scenarios with an average gap of 0.283. We further demonstrate ZAVIS with a Pioneer-3AT robot in real-world studies. 

\end{abstract}

\section{Introduction}

In this paper, we focus on the problem of efficiently locating a target object using a mobile robot equipped with vision sensors (e.g., an RGB-D camera). This task is often referred to as Active Visual Search (AVS)~\cite{99_ye} and is considered to be an essential component for successful social, service, or rescue robots. A robot's ability to search for a person or an object in a cluttered environment can be beneficial for many applications, including searching for a cellphone in an apartment for the owner, conducting a rescue mission to locate an injured person, removing hazardous materials at the scene of an accident, or even searching for evidence at the scene of a crime. 

Many of the existing work~\cite{19_zhang,20_zeng, 21_zheng} on AVS rely on the existence of a predefined set of object categories. However, this reliance places severe restrictions on AVS's applicability outside of the lab. Rather, real-world applications would be better served by AVS if it could accept object queries using \textit{free-form language} (e.g., "\texttt{red wallet}") and without dependence on predefined object categories. The task of searching for the target from a free-form language is often referred to as zero-shot object goal navigation~\cite{22_zhao, 22_gadre, 22_khandelwal}.

Similar tasks have been handled on the ALFRED challenge, which is also based on free-form language-based commands. The ALFRED (Action Learning From Realistic Environments and Directives) challenge~\cite{20_shridhar} is a set of robotics benchmarks that aims to solve a given task with a high-level language goal (e.g., rinse off a mug and place it in the coffee maker") and low-level instructions (e.g., "walk to the coffee maker on the right" and "wash the mug in the sink"), containing both navigation and object manipulation tasks. Then again, low-level instructions are costly as they rely on step-by-step human planning. In this work, we focus on active search of the target given as free-form language without any detailed instructions.

We propose an object search method that uses free-form language target commands, which we call Zero-shot Active VIsual Search (ZAVIS). ZAVIS can detect novel objects and plan search patterns to allow a robot to efficiently search for the target object based on a semantic grid map represented by static landmarks (e.g., desk or bed) while also using a commonsense-based knowledge prior encoded in language corpus~\cite{19_bosselut}. ZAVIS utilizes an open-set object detector to obtain the bounding box of unseen objects; this collection forms the set of candidate objects that are compared to the target object provided by a user in language form by using pre-trained vision-language models (i.e., CLIP~\cite{21_radford}). ZAVIS also includes a robot planning approach for object search: an efficient landmark-based planner equipped with a commonsense knowledge model to leverage practical prior knowledge~\cite{19_zhang}.
\section{Related work} \label{sec:related_work}

Active visual search (AVS) consists of a number of components including perception~\cite{18_redmon,15_ren} and planning~\cite{19_xiao, 96_ye}. The perception module (e.g., the detection or segmentation module) enables robots to detect and locate objects which can be either target objects or landmark objects. The planning module plans and obtains a high-level trajectory or waypoints that effectively search for the target object. 
In this work, we focus on both components in an open-set setting, where the label of the target object may not be included in the training datasets. Thus, in this section, we provide existing work on perception and planning algorithms for active search. 

Traditionally, AVS has been tackled with a detection module, and a planning module that work in a closed-set setting~\cite{21_zheng,19_zhang, 19_xiao, 96_ye}, where the label of the target object is included in the training dataset. Correlational Object Search (COS-POMDP)~\cite{21_zheng} formulates the task as a partially observable Markov decision process (POMDP) with a correlational observation model.
Unfortunately, this limits the target objects searchable by a robot. 
In order to overcome this problem, there has been an attempt to solve the zero-shot object goal navigation~\cite{22_gadre,22_khandelwal,22_zhao}. 
Gadre et al.~\cite{22_gadre} leverages Grad-CAM~\cite{16_selvaraju} of CLIP~\cite{21_radford} to locate the novel target objects and uses FBE~\cite{98_yamauchi} to explore to search for target objects. 
Our work stands along with these in that we also assume AVS in an open-set setting. 

\section{Problem Formulation}\label{sec:problem_formulation}

Zero-shot Active Visual Search describes the goal of searching for a target object that a user refers to in a free-form language. We call our task zero-shot since the target object is unseen during training and does not appear in predefined knowledge graphs. 
For this task, we assume that the location of the robot is accessible (i.e., GPS or wheel odometry), which is a common assumption adopted in other work~\cite{21_zheng,20_zeng,22_gadre}. The robot is also able to observe the world through RGB-D images and LiDAR sensor data. We further assume all of the objects in the scene are static, and the action space of the robot is continuous. 

The task is composed of two parts, planning and object detection, which are represented as modules in our system. Within the planning module, we define the target object as $o^{t}$ and the set of landmark names as $\mathcal{O}$. Landmark objects can be classified as either known $\mathcal{O}_k$ or unknown $\mathcal{O}_u$ during training while the target object is unknown. A set of N viewpoints of a landmark is denoted using the notation $\mathcal{P} = \{\mathbf{p}_i\}_{i=1}^N$ with $\mathbf{p}_i= \{o_i,x^v_i,y^v_i,\theta^v_i,I_i,\mathbf{b}_i\}$ where $o_i$ is a landmark object class, $x^v_i, y^v_i, \theta^v_i$ is a viewpoint position and orientation, $I_i$ is a raw input image and $\mathbf{b}_i$ is a bounding box from object detector corresponding to the object class $o_i$. On the detection module, our detection dataset consists of an input image $I$ and a set of bounding box and corresponding class $Y = \{c_i,\mathbf{b}_i\}_{i=1}^m$, where $m$ is a length of a set, $c_i$ is a class label and $\mathbf{b}_i$ is a corresponding bounding box. 
In addition, the cropped patch is denoted as $B = C(I,\mathbf{b})$, where $C$ is a crop function of an image $I$ and the bounding box $\mathbf{b}$.

%
%
\section{Proposed Method} \label{sec:proposed_method}

\begin{figure*}[!t]
    \centering
    \includegraphics[width=0.87\textwidth]{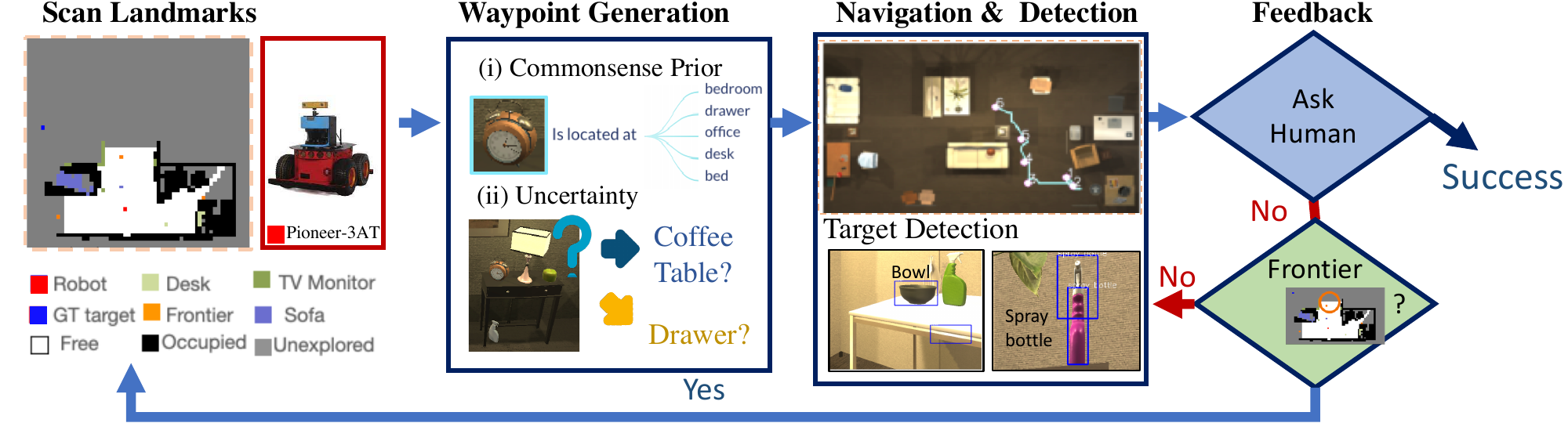}
    \caption{The overall procedure of ZAVIS. We present ZAVIS framework, object search based on \texttt{free-form language}. The framework consists of initial scanning, waypoint generation, navigation, and detection.
    The framework iteratively asks the human to compare the candidate patch set. }
    \label{fig:searching}
\end{figure*}

In this section, we discuss how our ZAVIS system tackles the active visual search problem. We explain the broad approach first, followed by details. 

\subsection{Zero-shot Active Visual Search}

ZAVIS takes an RGB-D image, LiDAR data, and the robot's location as inputs. In addition, the target object and the set of landmark object names are also given as inputs in the free-form language (presumably given by a user to the robot via an interface). The system's output is a candidate set of patches with a location projected on the map. The overall procedure of the algorithm is shown in Figure \ref{fig:searching}. 

As shown in Figure \ref{fig:searching} the algorithm starts with initial scanning for landmarks. The robot detects a landmark with the camera and extends the free space based on the LiDAR sensor. The images are obtained by rotating the pan-tilt camera, whose range follows the LiDAR sensor. For efficient search, target detection is also conducted concurrently during the initial scan. Lastly, the location of landmarks is projected to an internal map using depth information. 

The waypoint generator calculates the order of the visiting landmark, in which waypoint is a pose of the robot to view the landmark or explore. Waypoints are generated based on the map information, knowledge prior, and semantic detection uncertainty. We discuss further details of the waypoint generator in section \ref{sec:waypoint_generator}, and the method for obtaining the knowledge prior and semantic uncertainty in section \ref{sec:prior}.
Given the sequence of waypoints, the robot visits the landmarks and searches for the target object using the proposed object detection module and various viewpoints obtained by the pan-tilt camera. We defer the details of the detection module to the section \ref{sec:object_detection}. After successfully detecting a set of target candidate objects, the robot asks the human if the target is in the candidate set. If the target is in the candidate set, the search is complete. However, if the target is not found and the robot has visited all of the detected landmarks, the robot explores by moving to the nearest unexplored space, called the frontier. When the robot reaches the frontier, the robot reverts to the initial scanning phase and reiterates the procedure mentioned above.

\begin{figure}[!t]
    \centering
    \includegraphics[width=0.4\textwidth]{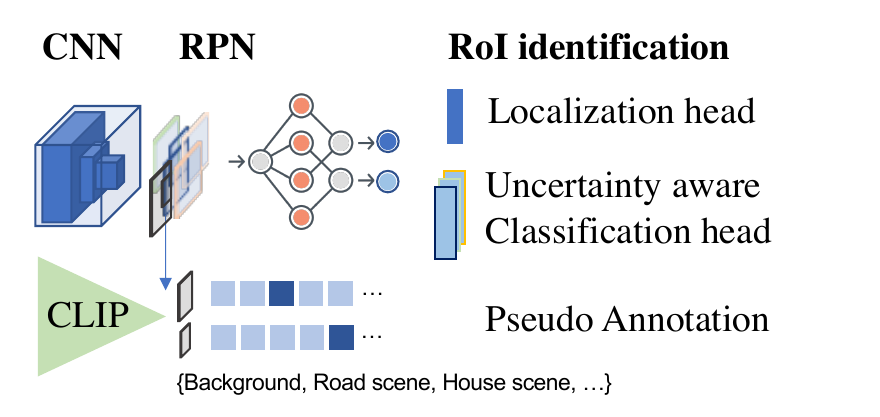}
    \caption{Proposed Open-set Object Detector. Our model classifies the unknown object by uncertainty-aware classification head, designed with a mixture of logit networks. Furthermore, we utilize the pretrained CLIP model to generate the pseudo annotation for the potential unknown objects.}
    \label{fig:my_label}
\end{figure}

\subsection{Object Detection} \label{sec:object_detection}

In order to search for objects given as free-form language, we also need to detect the objects whose label is not in the training dataset. We divide the problem into two stages: open-set object detection and text-image matching.

In the first stage, an open-set object detection approach is used with the aim of detecting both known and unknown labels trained only with a fixed set of labels. In this stage, the novel class is detected as a predefined '\texttt{unknown}' class. We follow the baseline model used in FasterRCNN~\cite{15_ren} with one additional class $K+1$ called '\texttt{unknown}'. Yet, we need to address two issues in order to detect novel objects: localizing the objects that are unknown and classifying these novel objects as members of the '\texttt{unknown}' class. For identifying the unseen object, we utilize the pretrained CLIP~\cite{21_radford} model to make a pseudo annotation for the potential unseen object. For classifying the novel object as '\texttt{unknown}', we also account for the uncertainty obtained by a mixture of logit networks. 

In order to localize the unknown object, we make pseudo annotations of the unknown class based on the RoI (Region of Interest) prediction of a Region Proposal Network (RPN)~\cite{15_ren}. We follow the prompt engineering setting \footnote{
The prompt set $\mathcal{S} = \{ \mbox{"a photo of "} o \mid o \in \{\texttt{"background"}, \\ \texttt{"road scene"}, \texttt{"house scene"}, \texttt{"animal"}, \texttt{"fashion accessory"}, \\
\texttt{"transport"}, \texttt{"traffic sign"}, \texttt{"home appliance"}, \texttt{"food"}, \\
\texttt{"sport equipment"}, \texttt{"furniture"}, \texttt{"office supplies"}, \\ \texttt{"electronics"},  \texttt{"kitchenware"}\} \}$
} of the CLIP model~\cite{21_radford} to measure the general objectness~\cite{18_redmon} of the RoIs. 
Next, we measure the CLIP objectness score by computing the distances between texts that correspond to the background (first three prompts from $\mathcal{S}$), then select the top-$k$ patches with higher CLIP objectness scores than a certain threshold, $0.9$. The CLIP objectness score of the patch $B$ is defined as follows:
\begin{equation}
    o(B) = 1- \sum_{s_i \in \mathcal{S}}^b \frac{f(B)\cdot g(s_i)}{\sum_{s_j \in \mathcal{S}} f(B)\cdot g(s_j) }
\end{equation}
where $b=3$ is a number of prompts corresponding to a background, $f$ is an image encoder and $g$ is a text encoder in CLIP\cite{21_radford}. We linearly increase the number of pseudo annotations for every iterations during training, which is denoted as $k$. In particular, we set $k = int(3*(iter/max \ iter))+1$. The pseudo annotation trains the whole network (i.e., RoI heads and RPN) for learning the unknown objects.

We use an uncertainty measure to categorize the novel object to '\texttt{unknown}' class. Although we have an additional class $K+1$ for unknown labels, we found it was not sufficient to capture all of the unknown objects. Rather, we utilize the mixture of logit network (MLN)~\cite{20_choi} on the RoI classification head to measure the uncertainty.
Due to the noisy pseudo annotation mentioned above, MLN is suitable as it can robustly learn the target. 

The output of the MLN head is composed of mixture logit $\boldsymbol{\mu}$ and mixture weight $\pi$. Then, the epistemic uncertainty is obtained by the disagreement of mixtures, and the aleatoric uncertainty is calculated by the weighted summation of entropy of the logits. Both epistemic uncertainty and aleatoric uncertainty is calculated as follows:
\begin{equation}
    \sigma_e^2 =\sum_{j=1}^J \left(\sum_{c=1}^{K+2}  \pi_j \left\| \mu^{(c)}_j-\sum_{m=1}^J \pi_m \mu^{(c)}_m\right\|^2 \right)
    \label{eq:epis}
\end{equation}

\begin{equation}
    \sigma_a^2 =\sum_{j=1}^J \pi_j \cdot \sum_{c=1}^{K+2}[-\mu_j^{(c)}\log\mu_j^{(c)} ] 
    \label{eq:alea}
\end{equation}
where $J$ is the number of mixtures, and $K+2$ is total number of the class labels including \texttt{background} and \texttt{unknown}. 

The loss function of the RoI classification head is as follows.
\begin{equation}
   \mathcal{L}_{cls} (\boldsymbol{\mu},\pi,c^*) = \sum_{j=1}^J \pi_j l({\boldsymbol{\mu}_j},c^*) \label{eq:MACE}
\end{equation}

where $l$ is a cross-entropy loss, and the ground-truth label is $\{c^*,\mathbf{b}^*\}$.


After training, we utilize the total uncertainty, i.e., the sum of epistemic and aleatoric uncertainty, to detect the unseen classes. We model the Gaussian distribution of total uncertainty on the training set; if the CDF of estimated total uncertainty is bigger than 0.9, we change the label of the patch to the 'unknown' class. 

The second stage in our object detection system is text-image matching. As CLIP embeds text and image features in the same space, we can use the distance between the embedding of text and image as a metric. We first make text embedding $g(o)$ with target object $o$, with the prompt mentioned in CLIP \texttt {"A photo of [object]"}. With open-set detection result $\{c_i,\mathbf{b}_i\}_{i=1}^n$, we crop the image based on bounding box corresponding to unknown class. We then measure the dot product of patches and text to gain the final matching score. The equation for the matching score is as follows:

\begin{equation}
    \label{eq:matching_score}
    m(o,I,\mathbf{b}_u) = f\left(C(I,\mathbf{b}_u)\right) \cdot g(o)
\end{equation}
where $f$ is the image encoder, $g$ is the text encoder, and $\mathbf{b}_u$ is an unknown class bounding box.

The detection module is used in both landmark detection and target object detection. In the case where detected landmarks belongs to a 'known class', we directly use the corresponding class $c_i$. If this is not the case, we crop the bounding box of '\texttt{unknown}' class and identify its class by conducting text-image matching with $\mathcal{O}_{u}$. In the case of target detection, as the target object assumes to be unseen during training, we conduct a text-image matching with bounding boxes whose label is '\texttt{unknown}'. If the matching score is bigger than a certain threshold $m_t$, we define the patch matches with the text.
%
%
\subsection{Waypoint Generator} \label{sec:waypoint_generator}

The robot searches for a target object by sequentially visiting waypoints which are generated based on the current map information and prior knowledge, (i.e., co-occurrence between landmark and target object). In order to obtain the waypoints, the viewpoints $\mathcal{P}$  for each landmark are first generated. The viewpoint is the position and orientation to view the landmark. 
Each viewpoint is calculated based on the gridmap $M$, with a predefined set of orientations. We utilize the heuristic that viewpoints must be located far enough to view the object and must be in the robot's reachable space. 



The waypoints is generated with a greedy algorithm based on the cost function. The cost function between current selected viewpoint $\mathbf{p}_c$ and next viewpoint candidate $\mathbf{p}_n$ is as follows.
\begin{multline}
    c(\mathbf{p}_{c},\mathbf{p}_{n}) = 
    \sqrt{(x_c-x_n)^2+ (y_c-y_n)^2} \\ +
    \lambda_1 (1+1^{-3}-p(o^{t}|o_n)))+ \lambda_2\sigma_{c}(I_n,\mathbf{b}_n)
    \label{eq:cost_function}
\end{multline}
where the $p(o^{t}|o_n)$ is co-occurrence measure of the landmark, $\lambda_1$ is a weight of a co-occurrence and $\lambda_2$ is a weight of a semantic uncertainty $\sigma_{c}(I_n,\mathbf{b}_n)$. Measurement of the co-occurrence and the semantic uncertainty is discussed in detail in section \ref{sec:prior}.

By the greedy algorithm, the candidate set of $\mathbf{p}_n$ is initialized as a set of viewpoints during scanning ($\mathcal{P}$), and the next viewpoint is selected, which has the minimum cost. If the co-occurrence of the viewpoint is less than threshold $t_c$ or the uncertainty is higher than $t_u$, the corresponding viewpoint is skipped. The viewpoint is removed from the candidate set after it is selected.

\subsection{Knowledge Prior \& Uncertainty} \label{sec:prior}

For efficient search, we also leverage prior knowledge, i.e., co-occurrence between query and landmark objects and the uncertainty of detected objects. Ambiguity in language is known to be inherent, leading to a semantic uncertainty of image-to-text matching. This ambiguity of text-image matching could lead to false text labels. For this reason, we add semantic uncertainty of unseen landmark objects to the cost function of the searching algorithm shown in Eq \ref{eq:cost_function}. 

For measuring the co-occurrence ($p(o^{t}|o_j, o_j \in \mathcal{O}_u$) between the landmark object and target object, we utilize the pre-trained commonsense knowledge model. Specifically, we use the commonsense transformer (COMET)~\cite{19_bosselut} BART trained on the ATMOIC 2020 dataset~\cite{21_hwang} to generate the possible locations where the target object would be located. We made an input of COMET $s_o$ as "\texttt{[Query name] [AtLocation] [GEN]}", distilling up to 20 generations from the COMET $\{G(s_o)_i\}_{i=1}^{20}$. Then, we estimate the maximum word similarity over the generated words and landmarks, which can be leveraged as a knowledge-based co-occurrence score.

\begin{equation}
    p(o^{t}|o) = \max_i [\ell\left(w(o^{t}), w(G(s_o)_i)\right)_{i=1}^{20}]
    \label{eq:co_occurrence}
\end{equation}
where $\ell$ denotes cosine similarity, and $w$ is for the word to vector embedding~\cite{13_mikolov}. 

To measure semantic uncertainty ($\sigma_c{p_{next}}$), we use the output entropy of a text-image matching model (i.e., CLIP). We define the probability that a certain patch of an image would belong to a specific landmark name $o_i$ as follows:
\begin{equation}
    p(o_i, I, \mathbf{b}) = \frac{\exp(m(o_i, I, \mathbf{b})/T_o)}{\sum_{j=1}^{L'} \exp(m(o_j, I, \mathbf{b})/T_o)}
\end{equation}
where $T_o$ is a temperature, and $o_i \in \mathcal{O}_u$. 
Then we calculate the entropy of the probability, defining it as a semantic uncertainty. 
\begin{equation}
    \sigma_c = -\sum_{i=1}^{L'} p(o_i, I, \mathbf{b})\log p(o_i, I, \mathbf{b})
    \label{eq:semantic_uncertainty}
\end{equation}

%
%
\section{Experiments} 
\label{sec:experiments}
In this section, we discuss the experimental results of the proposed method. We first validate the proposed open-set object detector on the vision dataset~\cite{10_everingham,14_lin}. We then quantitatively evaluate the proposed method in the simulation environment, AI2Thor simulation~\cite{17_kolve}. 
Finally, we qualitatively demonstrate the proposed method can be applicable to real-world scenarios and analyze the results.

\begin{table}[!t]
\caption{Openset Detection Result}
\centering
\resizebox{\columnwidth}{!}{%
\begin{tabular}{c||c|c|c|c}
     Method & WI & U-Recall & U- Precision & K-mAP \\
     \hline
     Faster RCNN~\cite{15_ren}& - & - & - & \textbf{56.35} \\
     ORE~\cite{21_joseph} & 0.04893 & 9.390 & 5.248 & 55.83 \\ 
     ZAVIS (Ours) &  \textbf{0.03999} & \textbf{19.06} & 2.279 & 53.67\\
\end{tabular}%
}
\label{tab:openset_det}
\end{table}

\subsection{OpenSet Object Detection}

\begin{figure}
    \centering
    \includegraphics[width=0.42\textwidth]{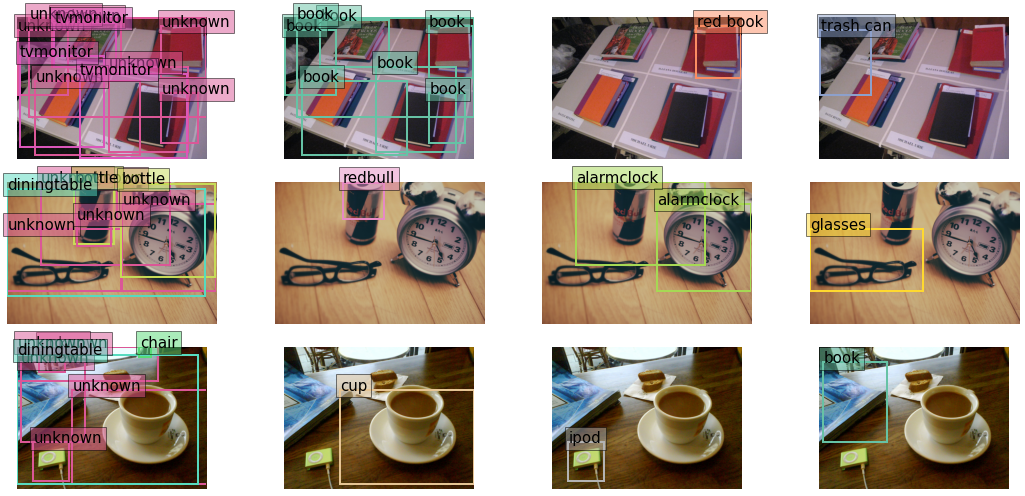}
    \caption{Predictions of proposed open-set object detector and text-image matching result. The first column is the result of open-set detection and the rest results from text-image matching. Note that 
    \texttt{book, red book, trash can, redbull, alarm clock, glasses, cup, ipod} is unseen during training and given as a language form in the text-image matching phase.}
    \label{fig:det_ex}
\end{figure}

\begin{table}[]
    \caption{Hyperparameters \& Object Settings}
    \resizebox{\columnwidth}{!}{%
    \begin{tabular}{c| c c c}
    \hline
    & RoboThor & IThor & RealWorld \\
    \hline
    $\lambda_1$ & 1 & 3 & 20\\
    $\lambda_2$ & 0.05 & 0.05 & 10 \\
    $t_c$ & 0.2 & 0.2 & 0.2 \\
    $t_u$ & 2.5 & 2.5 & 1 \\
    $m_t$ & 29 & 29 & 29\\
    \hline
    \multirow{4}{*}{target} & \multicolumn{2}{c}{RemoteControl, Laptop, Book} & Book, Cup, \\
    & \multicolumn{2}{c}{Apple, CD, Pot,} & Laptop,\\
    & \multicolumn{2}{c}{Bowl, AlarmClock,TeddyBear,} &  Cellphone\\
    & \multicolumn{2}{c}{CellPhone, SprayBottle, Pillow} & Tumbler \\
    \hline
    \multirow{2}{*}{Known Landmark} & \multicolumn{2}{c}{Tv monitor, Sofa,} &  Tv monitor\\
    & \multicolumn{2}{c}{Diningtable} & Diningtable \\
    \hline
    \multirow{3}{*}{Unknown Landmark} & \multicolumn{2}{c}{Armchair, Side table} & Table, Side table, \\
    & \multicolumn{2}{c}{Coffee table', Desk,} & Coffee Table, \\
    & \multicolumn{2}{c}{Bed, Drawer} & Desk\\
    \hline
    \end{tabular}
    }
    \label{tab:setting}
\end{table}

We validate the proposed open-set object detector with respect to wilderness impact (WI)~\cite{20_dhamija}, recall of unknown class (U-Recall), the precision of the unknown class (U-Precision), and mean average precision of the known class (K-mAP). WI measures how much the model is confused with unknown and known labels, and U-Recall measures how much the detectors capture the unknown objects with respect to unknown objects in the test set. 
All of the detectors are trained on the Pascal VOC 2012 dataset~\cite{10_everingham} and the subset MS COCO dataset~\cite{14_lin} that has the same class label as the Pascal VOC dataset. The result are shown in Table \ref{tab:openset_det}, and the qualitative results are in Figure \ref{fig:det_ex} . 

As the baseline Faster RCNN~\cite{15_ren} is not designed for the open-set detection, we can not measure any metric related to the open-set setting. Compared with an open-set detector ORE~\cite{21_joseph}, our proposed detector has a lower WI $0.00894$ and higher U-recall $9.67$.
Although the U-Precision of our method is lower, we observe that the model with higher recall better fits on object searching problem. Because the algorithm is designed to ask for humans after target detection, the success depends on the ability to capture the target object regarding false prediction.

\subsection{Simulational Result}

\begin{figure}
    \centering
    \includegraphics[width=0.42\textwidth]{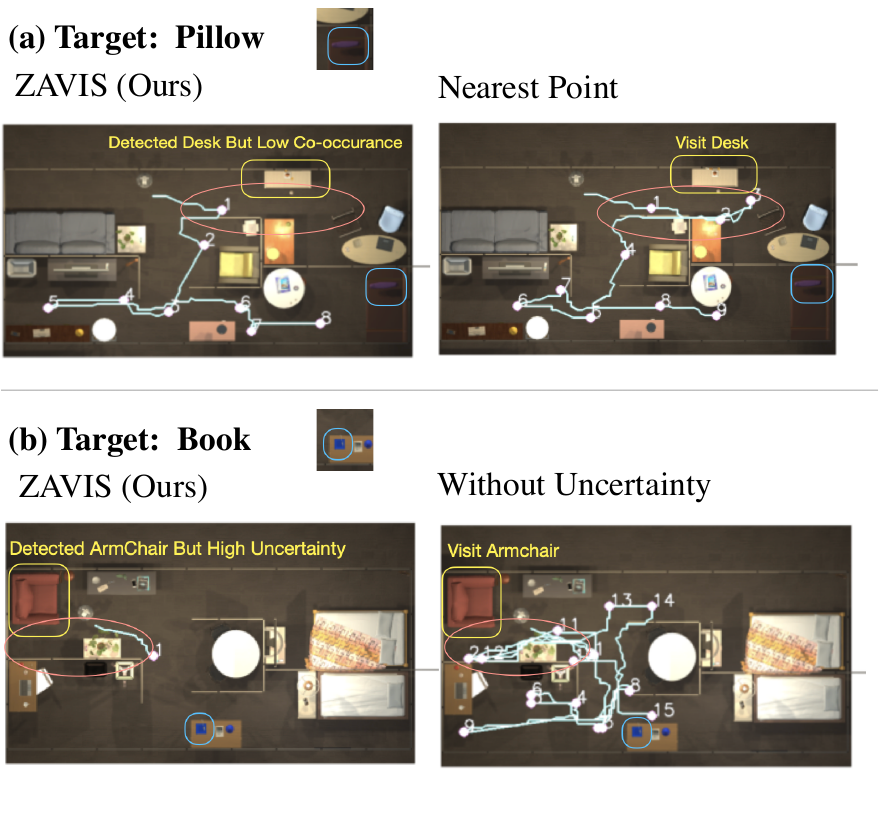}
    \caption{Examples of trajectory generated in ablation studies on AI2THOR. \textit{(Top)} compares the path between studies without the co-occurrence module. \textit{Bottom} compares the path between other methods without considering the uncertainty. }
    \label{fig:sim_demo}
\end{figure}

\begin{table}[!t]
\caption{Comparison with Previous Work}
\centering
\begin{tabular}{c c||c|c}
     scene & method & SPL & SR (\%) \\
     \hline
     \multirow{3}{*}{RoboThor} 
     & COS-POMDP~\cite{21_zheng} \textdagger & 0.0693 & 14.38 \\
     & CoW~\cite{22_gadre} & 0.1154 & 51.91 \\
     & ZAVIS (Ours) & \textbf{0.3462} & \textbf{91.80} \\
     \hline
     \multirow{3}{*}{IThor-Bedroom} 
     & COS-POMDP~\cite{21_zheng} \textdagger & 0.2334 & 49.21 \\
     & CoW~\cite{22_gadre} &  0.093 & 41.09 \\
     & ZAVIS (Ours) & \textbf{0.3017} & \textbf{81.42} \\
     \hline
     \multirow{3}{*}{IThor-Livingroom} 
     & COS-POMDP~\cite{21_zheng} \textdagger & 0.1410 & 29.16 \\
     & CoW~\cite{22_gadre} & 0.00088 & 14.70 \\
     & ZAVIS (Ours) & \textbf{0.1969} & \textbf{75.00}\\
     \hline
\end{tabular}
\label{tab:comparison}
\end{table}

\begin{table}[t]
\caption{Ablation Studies}
\centering
\begin{tabular}{p{0.17\linewidth} p{0.3\linewidth} || p{0.1\linewidth} | p{0.1\linewidth} | p{0.1\linewidth}}
     Ablations & & SPL & SR (\%) & \# \\
     \hline
     \multirow{2}{*}{Detectors} & FasterRCNN~\cite{15_ren} & 0.2600 & 75.95 & 9.48\\
     & ORE~\cite{21_joseph} & 0.2835 & 85.79 & 7.71 \\
     \hline
     \multirow{3}{*}{Co-occurrence} & 
     Nearest Point & 0.3293 & 89.61 & 7.05 \\
     & Web-based~\cite{20_zeng}& 0.3210 & 89.00 & 7.36 \\
     & Word Similarity~\cite{13_mikolov} & 0.3374 & 90.71 & \textbf{6.56} \\
     \hline
     \multirow{1}{*}{Uncertainty} &
     w/o Uncertainty & 0.3292 & 87.91 & 8.66 \\
    \hline 
     & ZAVIS (Ours) & \textbf{0.3462} & \textbf{91.80} & 6.65\\
     \hline 
\end{tabular}
\label{tab:ablation}
\end{table}

\begin{figure*}[!t]
    \centering
    \includegraphics[width=0.96\textwidth]{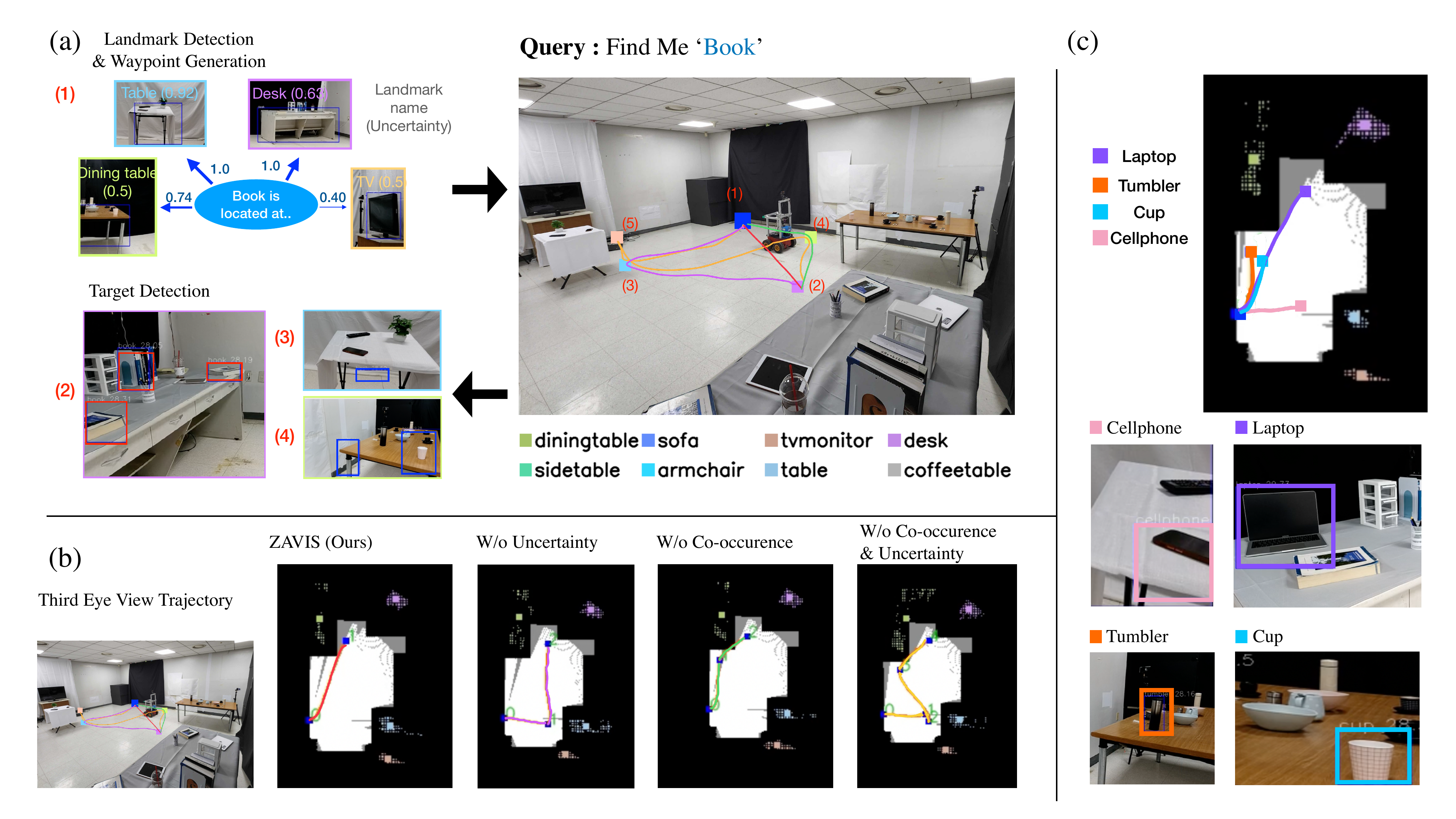}
    \caption{Real-world Demonstration. 
    (a) Explains the demonstration on the target \texttt{book}. 
    (b) shows the trajectory for four scenarios, the proposed method, without semantic uncertainty, without co-occurrence, and without both.
    (c) shows a various trajectory and detection result with target \texttt{book, notebook, tumbler, cup, cellphone}}
    \label{fig:realword}
\end{figure*}

We conducted experiments of ZAVIS in simulated 3D scenes using the IThor and RoboThor~\cite{20_deitke} environments within the open source AI2Thor framework, a tool for visual AI research~\cite{17_kolve}. Using the IThor environment, we validate the proposed method in multiple bedrooms and living room scene types, each having 30 scenes. Additionally, we also use a total of 15 scenes from the RoboThor environment to help validate ZAVIS. 
In our experiments, however, we mainly use RoboThor as the environment has diverse multi-room scenes, whereas IThor is limited to containing single or double rooms. 
We use an RRT planner~\cite{01_lavalle} for the navigation, and the hyperparameters and object settings are described in Table \ref{tab:setting}.

As the proposed method asks humans whether the target is in the image, success is defined as the condition where the candidate set contains the ground-truth target object in the simulated environment. Particularly, We denote the target is detected if the IoU (intersection over union) between ground-truth objects bounding box and the candidate bounding box is bigger than 0.3 or IoA (intersection over area) bigger than 0.5. We define a failing condition to be when no patch corresponds to the ground truth image while the robot is moving more than 50m. 


We evaluate the proposed method with two commonly used evaluation metrics for active visual search, SPL (Success weighted by Path Length)~\cite{18_anderson} and SR (success rate). We benchmark two methods, COS-POMDP~\cite{21_zheng} and CoW~\cite{22_gadre}. As COS-POMDP~\cite{21_zheng} does not fit for zero-shot setting on both detector and correlation model, we changed the detector to Faster RCNN with the CLIP model and web-based correlation model \cite{21_zheng}. 

The experimental results on the simulation environment are shown in Table \ref{tab:comparison}. The method with \textdagger is adjusted for the zero-shot setting. The proposed method outperforms those of all previous work by a significant margin with an average gap $0.173$ in SPL and $49.33\%$ in SR.
Thus, we observe that the proposed detection method can locate unseen objects better than the Grad-CAM-based method based on success rate. 
In addition, while CoW~\cite{22_gadre} directly searches for the objects based on FBE~\cite{98_yamauchi} without any landmarks, our proposed ZAVIS method searches for both target objects and co-occurrences with landmark objects for efficient search. 
We have also observed that the proposed method outperforms algorithms designed for non-zero-shot AVS with the adjustment to the zero-shot setting. 

We further conduct the ablation studies to show the effect of the proposed detector, co-occurrence, and uncertainty. We experimented on various object detectors, Faster-RCNN~\cite{15_ren}~\cite{21_radford} and ORE~\cite{21_joseph}. In addition, we evaluate the various co-occurrence measure, web-based knowledge prior~\cite{20_zeng}, word similarity~\cite{13_mikolov} and without any co-occurrence measure.
The results are summarized in Table \ref{tab:ablation}, where \# denotes the average number of waypoints.
Figure \ref{fig:sim_demo} shows an example of the trajectory. The robot with target as \texttt{pillow} skips to visit the \texttt{desk} as it has low co-occurrence, while the method that does not utilizes co-occurrence visit the \texttt{desk} as shown in (a). In addition, in the case of  the \texttt{pillow} target (b), the robot does not visit the \texttt{armchair} due to high uncertainty, inducing an efficient trajectory.

%
%
\subsection{Real-world Demonstration}

In this section, we discuss the qualitative results of a real-world pilot experiment where ZAVIS was implemented on a robot. We used the Pioneer-3AT robot with a pan-tilt stereo camera and LiDAR sensor. A DWA planner (dynamic window approach)~\cite{97_fox} is implemented for the local planning phase to consider the motion dynamics. 
The target and landmark objects are shown in Table \ref{tab:setting}. 
We carefully placed the objects at or near landmarks based on the feasibility of the object/landmark co-occurrence (e.g., book on the desk, cup on the dining table).

Figure \ref{fig:realword}-(a) shows the demonstration in which robot is searching for the \texttt{book}. The robot looks for the landmark and target in the first phase and successfully finds and locates all landmark objects. 
The robot plans to visit the desk first because the desk has a high co-occurrence score and low semantic uncertainty. Although the table has high co-occurrence and is closer, the robot decides to visit the table after the desk due to high uncertainty. After the decision, the robot moves to the landmark \texttt{desk} and detects the book, and asks the human if the book exists in the detected patch set. 

We then show the effect of the proposed co-occurrence measure and semantic uncertainty measure with the target \texttt{book} shown in Figure \ref{fig:realword}-(b). Without the uncertainty measure, the robot visits closer landmarks with high co-occurrence \texttt{table} first, then visit the \texttt{desk}. Without a co-occurrence measure, the robot plans to visit \texttt{dining table} first as it has low uncertainty and is close to its current position. Without both of these measures, the robot only plans based on the distance, leading to the longest trajectory. In addition, we demonstrate the proposed method on various target objects shown in Figure \ref{fig:realword}-(c). 

%
%
\section{Conclusion}

In this paper, we have introduced a system that enables a robot to actively move around the environment to search and locate the target object that are unknown to the robot and proposed Zero-shot Active Visual Search (ZAVIS). To this end, we have presented three novel modules for successful zero-shot active visual search. Firstly, we have introduced a novel open-set object detector that can detect unseen objects during training and expand it for zero-shot detection with a text-image matching model. Secondly, we utilized the commonsense-based language model to gain prior knowledge regarding co-occurrences between landmark and target objects. We believe that a large language model with commonsense can better generalize the co-occurrence as it contains coordinating and relationship knowledge shared amongst people. Lastly, we have shown that considering the uncertainty from the text-image matching model helps efficient search. As part of this study, we have validated the effectiveness and utility of ZAVIS during experiments performed in both simulated and real-world environments. 


\bibliographystyle{unsrt} 
 \bibliography{references}
\end{document}